\documentclass[journal]{IEEEtran}
\usepackage[pdftex]{graphicx}
\graphicspath{{./images/}{./plots/}{./sketches/}}
\DeclareGraphicsExtensions{.pdf,.jpeg,.png}
\usepackage{bm}
\usepackage{algorithm}
\usepackage{algorithmicx}
\usepackage{algpseudocode}
\usepackage[cmex10]{amsmath}
\usepackage{amssymb}

\usepackage[caption=false,font=footnotesize]{subfig}

\usepackage{fixltx2e}

\usepackage{stfloats}
\fnbelowfloat

\begin{document}
%
\title{Passive Reaction Analysis for Grasp Stability}
%
%
%

\author{Maximilian~Haas-Heger,~\IEEEmembership{Student Member,~IEEE,}
        Garud~Iyengar,~\IEEEmembership{Member,~IEEE,}
        and~Matei~Ciocarlie,~\IEEEmembership{Member,~IEEE,}
\thanks{M. Haas-Heger and M. Ciocarlie are with the Department
of Mechanical Engineering, Columbia University, New York,
NY, 10027 USA (email: m.haas@columbia.edu, matei.ciocarlie@columbia.edu).}
\thanks{G. Iyengar is with the Department of Industrial Engineering 
and Operations Research, Columbia University, New York, NY, 10027 USA (email: garud@ieor.columbia.edu).}
\thanks{Manuscript received May 5, 2017. This work was supported in part by the Office 
of Naval Research Young Investigator Program under award 
N00014-16-1-2026.}}

%
%

\markboth{Transactions on Automation Science and Engineering}%
{Haas-Heger \MakeLowercase{\textit{et al.}}: On the Distinction between Active and Passive Reaction in Grasp Stability Analysis}


\maketitle

\begin{abstract} In this paper we focus on the following problem in multi-fingered
robotic grasping: assuming that an external wrench is being applied to
a grasped object, will the contact forces between the hand and the
object, as well as the hand joints, respond in such a way as to
preserve quasi-static equilibrium? In particular, we assume that there
is no change in the joint torques being actively exerted by the
motors; any change in contact forces and joint torques is due
exclusively to passive effects arising in response to the external
disturbance. Such passive effects include for example joints that are
driven by highly geared motors (a common occurence in practice) and
thus do not back drive in response to external torques. To account for
non-linear phenomena encountered in such cases, and which existing
methods do not consider, we formulate the problem as a mixed integer
program used in the inner loop of an iterative solver. We present
evidence showing that this formulation captures important effects for
assessing the stability of a grasp employing some of the most commonly
used actuation mechanisms.

\textit{Note to practitioners:} Once a grasp of a given object has
been chosen, our method has multiple possible applications. First, it
can be used to determine how the choice of a pre-load (i.e. the
torques applied to the joints as the grasp is created) affects the
stability of the grasp. Second, once a pre-load has been chosen, our
method can be used to determine which external disturbances can be
resisted solely through passive effects, without further changes to
the commands sent to the motors. We believe this approach is
particularly relevant for the large family of grasping devices that
are not equipped with tactile or proprioceptive sensors, and are thus
unable to sense external disturbances or to control joint torques, but
are still effective thanks to passive resistance effects.

\end{abstract}


\section{Introduction}

\IEEEPARstart{S}{tability} analysis is one of the foundational problems for
multi-fingered robotic manipulation. Grasp planning, or the problem of
determining an appropriate grasp for a given object using a particular
hand design,  
can be posed as a search over the space of possible
grasps looking for instances that satisfy a measure of stability. The
formulation and characteristics of the stability measure thus play a
key role in this search, and, by extension, in any task that begins by
planning and executing a grasp.

In turn, multi-fingered grasp stability relies on studying the net
resultant wrench imparted by the hand to the grasped object. Ferrari
and Canny~\cite{FERRARI92} introduced a very efficient geometric
method for determining the total space of possible resultant wrenches
as long as each individual contact wrench obeys (linearized) friction
constraints. This method answers the simplest form of what we refer to
here as the existence problem: given a desired output, are there legal
contact wrenches that achieve it, and, if so, how large is their
needed magnitude in relation to the output? This approach has been at
the foundation of numerous planning algorithms proposed since. 

Consider a grasp that scores highly according to the quality metric
described above. This means that any desired resultant can be
produced by a computable set of contact wrenches (of bounded
magnitude). In turn, the contact wrenches can be balanced by a set of
joint torques, which can also be computed~\cite{handbook2008}. 
However, this approach is based on a string of assumptions:
\begin{itemize}
\item First, we have assumed that at every contact we can actively 
control the forces exerted.
\item Second, we have assumed that, at any given moment, the control
  mechanism knows what resultant wrench is needed on the object.
\item Third, we have assumed that the joint torques needed to balance
  this resultant can be actively commanded by the motor outputs.
\item Fourth and finally, we have assumed that the desired motor output
  torques can be obtained accurately.
\end{itemize}

In practice, these assumptions do not always hold. First, many robotic
hands contain members with limited mobility. That is, the contact
forces  on any such member cannot be arbitrarily controlled. This is
of particular importance when enveloping 'power' grasps are
considered, where contact between the object is not only made at the
fingertips, but also at the proximal links and the palm of the hand.
The force at a contact can only directly be affected through the actuators
preceding the contact in the kinematic chain. Hence, for a contact on the palm the forces cannot be directly
controlled using finger actuators. Instead, the force at such a
contact will arise indirectly due to the externally
applied wrench and the contact forces exerted at other contacts.

Second, the external wrench in need of balancing is difficult to
obtain: to account for gravity and inertia, one needs the exact mass
properties and overall trajectory of the object, which are not always
available; any additional external disturbance will be completely
unknown, unless the hand is equipped with tactile sensors. Third, the
kinematics of the hand may not permit explicit control of joint
torques. This is of particular importance for the large class of
underactuated hands, as by definition, the joint torques may not be
independently controlled. Instead they are determined by the kinematic
structure of the hand.  Finally, many commonly used robot hands use
highly geared motors unable  to provide accurate torque sensing or
regulation.

A much simpler approach to establishing a stable grasp, applicable to 
more types of hardware, is to simply select a set of motor commands 
that generate some level of \textit{preload} on the grasp, and maintain 
that throughout the task. This method assumes that the chosen motor 
commands will not only lead to an adequate and stable preload for grasp 
creation, but also prove suitable for the remainder of the task. A key 
factor that allows this approach to succeed is \textit{the ability of a grasp to absorb
  forces that would otherwise unbalance the system without requiring
  active change of the motor commands.}

Following the arguments above, we believe it is important to not only
consider the wrenches the hand can apply actively by means of its
actuators, but also the reactions that arise passively. Thus, in this
study we are interested in the distinction between \textit{active
force generation}, directly resulting from forces applied by a
motor, and \textit{passive force resistance}, arising in response to
forces external to the contacts or joints. Consider the family of highly
geared, non-backdrivable motors: the torque applied at a joint can
exceed the value provided by the motor, as long as it arises
passively, in response to torques applied by other joints, or to
external forces acting on the object. Put another way, joint torques
are not always the same as output motor torques, even for direct-drive
hands.

The distinction between active and passive force resistance is also an 
important feature of underactuated hands. Wrench resistance is highly 
reliant on passive compliance, and contact forces in the nullspace of 
the transposed grasp Jacobian are common due to the underactuated 
nature of the hand. These contact forces will have no effect on the 
actuator but cause a purely passive reaction. The existing grasp 
stability analysis tools are not equipped to account for the behavior
of non-backdrivable actuators or underactuated kinematics. 

From a practical standpoint, a positive answer to the existence
problem outlined above is not useful as long as the joint torques
necessary for equilibrium will not be obtained given a particular set
of commands sent to the motors. Here, we focus on stability from an inverse
perspective: given a set of commanded torques to be actively applied
to the robot's joints, what is the net effect expected on the grasped
object, accounting for passive reactions?

Overall, the main contribution of this paper is a quasi-static
grasp stability analysis framework to determine the \textit{passive} response
of the hand-object system to applied joint torques and externally
applied forces. This method was designed to account for actuation
mechanisms such as non-backdrivable or position-controlled motors
without explicit torque regulation. Furthermore, it enables the 
analysis of the passive behavior of some of the most commonly used 
underactuated hand mechanisms.

An initial report from this work appeared in the 2016 Workshop on the
Algorithmic Foundations of Robotics. This study extends on the
previous report in multiple ways. We improve the discussion of the
underlying physics of the problem, including the aspects that can not
be handled exactly and require approximation in our approach. We
extend the formulation to account for tendon-driven underactuated
hands (which proves to be a generalization of the per-joint treatment
shown previously), and present additional experiments to validate the
applicability of our method.

\section{Related Work}

The problem of force distribution between an actively controlled robotic 
hand and a rigid object has been considered by a number of authors~\cite{
SALISBURY83,AICARDI96,KERR86,YOSHIKAWA91}. A great simplification to grasp 
analysis is the assumption that any contact force can be applied by 
commanding the joint motors accordingly. This assumption neglects the 
deficiency of the kinematics of many commonly used robotic hands in 
creating arbitrary contact forces. The idea that the analysis of a grasp 
must include not only the geometry of the grasp but also the kinematics of 
the hand is central to this paper.

Bicchi~\cite{BICCHI93} showed that for a kinematically deficient hand only 
a subset of the internal grasping forces is actively controllable. Using a 
quasi-static model, the subspace of internal forces was decomposed into 
subspaces of active and passive internal forces. Making use of this 
decomposition Bicchi proposed a quadratic program formulation for the problem of optimal 
distribution of contact forces with respect to power consumption and given 
an externally applied wrench~\cite{BICCHI94}. He 
proposed a definition of force-closure that makes further use of this 
decomposition and developed a quantitative grasp quality metric that 
reflects on how close a grasp is to losing force-closure under this 
definition~\cite{BICCHI95}. 

Prattichizzo et al.~\cite{PRATTICHIZZO97} made 
use of the previous work by Bicchi to compute the subset of actively 
controllable internal contact forces and proposed two grasp quality 
measures that are applicable to kinematically deficient grasps. They
analyze how far a set of contact forces are from violating their contact
constraints and derive a potential contact robustness (PCR). Any wrench 
with magnitude less or equal to this parameter can be reacted without violation 
of these constraints. They furthermore define a potential grasp robustness
(PGR), which is similar to the PCR, but allows for contacts either breaking
or sliding. Both PCR and PGR are conservative metrics and the calculation of
the PGR can be computationally infeasible at higher number of contacts due to
its combinatorial nature. Prattichizzo et
al.~\cite{PRATTICHIZZO13} also investigated the controllability of object
motion and contact forces in underactuated hands using a quasi-static
grasp model. Specifically, they consider compliant hands that exhibit
passive mechanical adaptation and make use of ``postural synergies''.

There have been rigid body approaches to the analysis of active and passive 
grasp forces. Yoshikawa~\cite{YOSHIKAWA96} studied the concept of active 
and passive closures and the conditions for these to hold. Melchiorri~\cite{
MELCHIORRI97} decomposed contact forces into four subspaces using a rigid 
body approach. Burdick and Rimon~\cite{RIMON16} formally defined four 
subspaces of contact forces and gave physically meaningful interpretations. 
They analyzed active forces in terms of the injectivity of the transposed 
hand Jacobian matrix. They note that the rigid body modeling approach is a 
limitation, as a compliance model is required to draw conclusions on the 
stability of a grasp. 

An important distinction between our work and that of the above authors 
lies in the definition of what qualifies as a ``passive'' contact force. In 
addition to contact forces that lie in the null space of the transposed 
hand Jacobian, we consider contact forces arising from joints being loaded 
passively (due to the non-backdrivability of highly geared motors), and not 
arising from the commanded joint torque, as passive. Furthermore, we define 
preload forces as the internal forces that arise from selecting a set of 
motor commands that achieve a grasp in stable equilibrium, previous to the 
application of any external wrench.

\section{Problem Statement}

Consider a grasp establishing multiple contacts between the robot hand
and the grasped object. We denote the vector of contact wrenches by
$\bm{c}$. In equilibrium, the grasp map matrix $\bm{G}$ relates contact
wrenches to the wrench applied to the object externally $\bm{w}$,
while the transpose of the grasp Jacobian $\bm{J}$ relates contact
wrenches to joint torques $\bm{\tau}$:
\begin{eqnarray}
\bm{G} \bm{c} &=& -\bm{w} \label{eq:wrench}\\
\bm{J}^T \bm{c} &=& \bm{\tau} \label{eq:tau}
\end{eqnarray}

One way to apply this in practice is to check, for a given wrench
$\bm{w}$, if contact forces exist that satisfy (\ref{eq:wrench}). For
a comparison of different methods that can be used to compute contact
forces, that satisfy this and other constraints, see the recent work
by Cloutier et al~\cite{CLOUTIER18}. In order to also consider the
hand mechanism, one could then also check if joint torques exist that
further satisfy (\ref{eq:tau}). However, as we discuss below, this
application method has important shortcomings.

Assume that, for given wrench $\bm{w}$, a given set of contact forces
and joint torques have been found to satisfy the equilibrium
conditions above; denote these by $\bm{c}_{eq}$ and $\bm{\tau}_{eq}$
respectively. The most straightforward way to use this would be to command
the motors to achieve $\bm{\tau_{eq}}$; in other words, if
$\bm{\tau_c}$ is the command sent to the motors, we simply ask that
$\bm{\tau}_c = \bm{\tau}_{eq}$. However, this approach is subject to
the assumptions outlined in the Introduction: it requires that
$\bm{w}$ be known, that applying $\bm{\tau}_{eq}$ at the joints in the
presence of $\bm{w}$ actually results in the desired contact forces
$\bm{c}_{eq}$, and finally that we can produce desired joint torques
$\bm{\tau}_{eq}$. The final point further implies that the hand has
the needed control authority over all needed degrees of freedom (which
is rare, since most fingers are kinematically deficient with at most 3
joints), and that motors can regulate their torque output as needed
(also rare, as most hands used in practice are position- and not
force-controlled).

A much more common approach is to command a given $\bm{\tau}_c$, and
rely on $\bm{\tau}_{eq} \neq \bm{\tau}_c$ \textit{arising through
  passive reactions}. For the large family of hands powered by geared,
non-backdrivable motors, at any joint $i$ the resulting torque
$\tau_i$ can exceed the commanded value, but only passively, in
response to the torques $\tau_j,~j \neq i$ and the wrench $\bm{w}$. If
this happens as desired for a range of disturbances $\bm{w}$, then the
job of controlling the hand is greatly simplified: one just needs to
always command $\bm{\tau}_c$, and the reaction that stabilizes any
particular $\bm{w}$ happens passively. However, the field currently
lacks a method to accurately analyze this problem.

We state our problem as follows: \textit{given commanded torques
  $\bm{\tau}_c$, can the system find quasi-static equilibrium for a
  disturbance wrench $\bm{w}$, assuming passive reaction effects at the
  joints?}

\subsection{The classical approach} 

In combination with a contact constraint model, solving the relatively
simple system of Eqs. (\ref{eq:wrench})\&(\ref{eq:tau}) for $\bm{c}$
and $\bm{\tau}$ when $\bm{w}$ is given amounts to solving a force
distribution problem. However, for rigid bodies, this problem of
computing the exact force distribution across contacts in response to
an applied wrench is statically indeterminate.

Previous studies such as those of
Bicchi~\cite{BICCHI93,BICCHI94,BICCHI95} make use of a linear
compliance matrix that characterizes the elastic elements in a grasp
and solves the indeterminacy. For a comprehensive study on how to
compute such a compliance matrix see the works by Cutkosky and
Kao~\cite{CUTKOSKY_COMPLIANCE}, and Malvezzi et al. for an extension to underactuated hands~\cite{MALVEZZI13}. A compliance matrix allows us to
consider the force distribution across contacts as the sum of a
particular and a homogeneous solution. The contact forces $\bm{c_0}$
create purely internal forces and hold the object in equilibrium. This
is the homogeneous solution and, as noted in the Introduction, it can
be of great importance to grasp stability. The contact forces
$\bm{c_p}$ associated with the application of an external wrench
$\bm{w}$ are considered the particular solution. Bicchi formulates a
\textit{force distribution problem}~\cite{BICCHI94} given by $\bm{c} =
\bm{c_p} + \bm{c_0} = \bm{G^R_K}\bm{w} + \bm{c_0}$ where $\bm{G^R_K}$
is the $\bm{K}$-weighted pseudoinverse of the grasp map matrix
$\bm{G}$. $\bm{K}$ is the stiffness matrix of the grasp and is given
by the inverse of the grasp compliance matrix.
\begin{figure*}[!t]
\centerline{\subfloat[Homogeneous]{\includegraphics[width=2.5in]{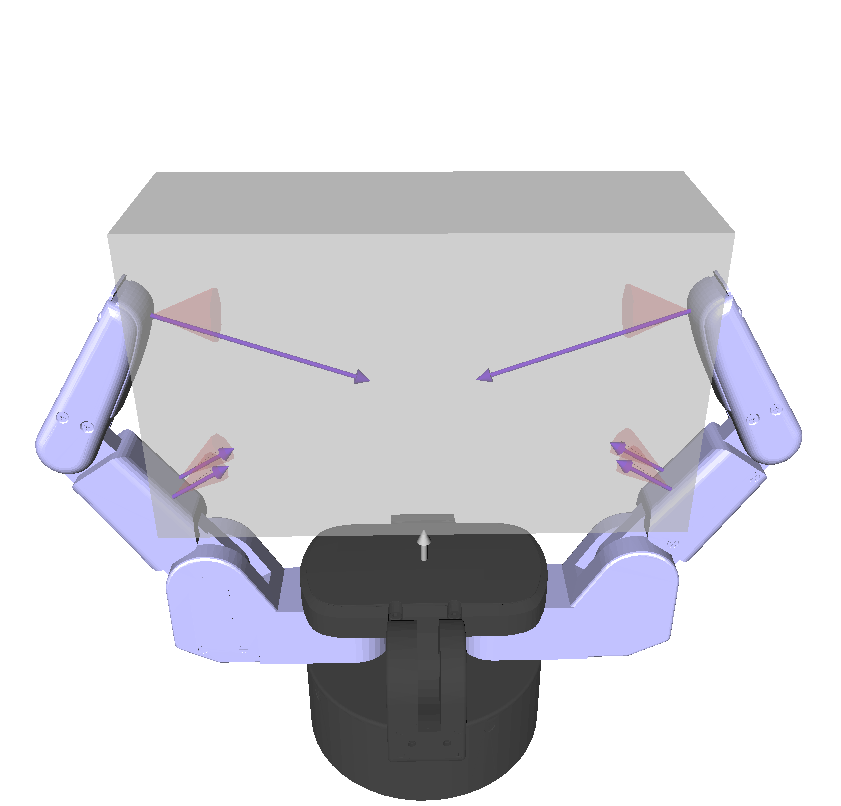}%
\label{fig:compliance_preload}}
\hfil
\subfloat[Homogeneous + Particular]{\includegraphics[width=2.5in]{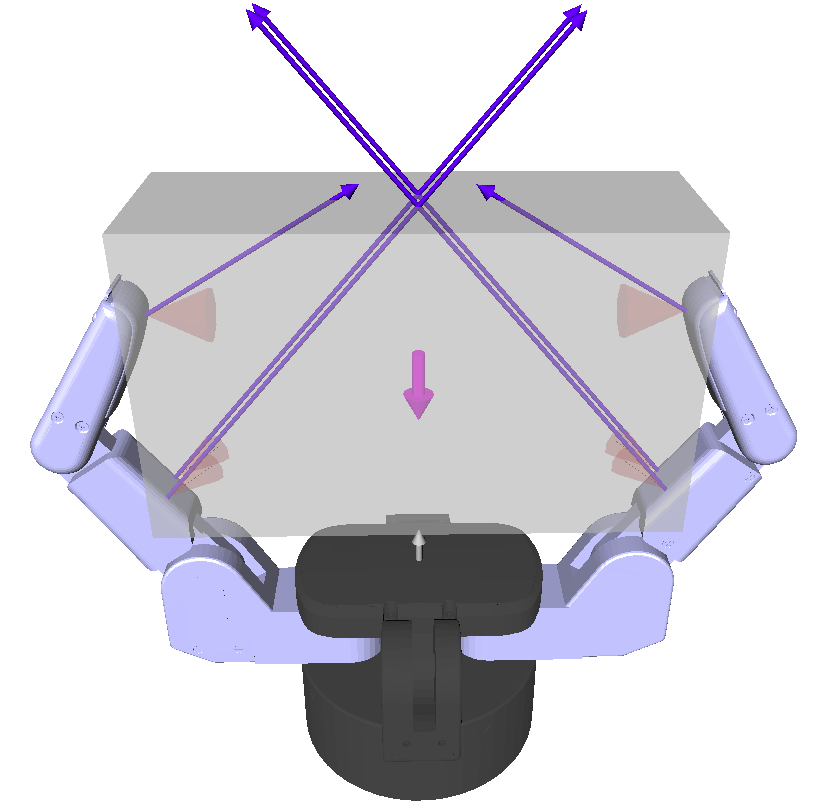}%
\label{fig:compliance_wrench}}}
\caption{Illustration of the shortcomings of a linear compliance model. The homogeneous solution was obtained using the algorithm presented in this paper. The particular solution was computed using the linear compliance approach~\cite{BICCHI94}. Contacts have unity stiffness in the normal direction. The stiffness in the frictional direction was set equal to the coefficient of friction. The joints are assumed infinitely stiff. Friction cones are shown in red and corresponding contact forces are shown as blue arrows. The violet arrow denotes the applied force.}
\label{fig:compliance}
\end{figure*}

Given the subspace of controllable internal forces~\cite{BICCHI93}, the particular solution computed in this way can be used to compute a homogeneous solution such that $\bm{c}$ satisfies all contact constraints. Using Eq. (\ref{eq:tau}) the required equilibrium joint torques that satisfy this
system $\tau_{eq}$ can then be calculated.

\subsection{Limitations of this approach} 

The use of a linear compliance matrix is an important limitation, as
it assumes a linear stiffness of the contacts and the joints. However,
a contact force may only ``push'', it cannot ``pull'' and hence
contact forces behave in a nonlinear fashion. Furthermore, a linear
compliance model disregards the nonlinearity of frictional forces
obeying Coulombs law of friction. We consider contacts of the
\textit{point contact with friction} type, which means that the
contact force must lie within its friction cone. A linear compliance
model has no notion of this friction constraint and thus cannot
distribute forces accordingly once the frictional component of a
contact force reaches its limit.

To illustrate this issue, consider
Fig.~\ref{fig:compliance}. Fig.~\ref{fig:compliance_preload} shows a
homogeneous solution to a force distribution
problem. Fig.~\ref{fig:compliance_wrench} shows the sum of the
homogeneous and particular solutions when an external wrench pushing
the object towards the palm of the robot is applied. Applying a
downward force has caused the contact forces on the distal links to
violate the friction constraint (they lie outside their respective
cones), perhaps leading us to believe that we have to increase the
internal forces in the grasp in order to resist the applied wrench. In
reality, however, the contacts on the distal links will only apply as
much frictional force as they may, and more force will be distributed
to the contacts on the proximal links. Indeed, experimental results
indicate that this grasp withstands arbitrary downward forces applied
to the object even in the absence of internal forces.

An attempt to alleviate this issue has been made by Prattichizzo et
al. in their work on the PGR quality
measure~\cite{PRATTICHIZZO97}. The computation of this metric allows
for a contact to slide or break entirely. In order to linearize the
problem, however, a sliding contact may not exert any frictional
forces at all. Thus, the contact forces obtained through this method
may be not be physically motivated and the PGR measure tends to be a
conservative quality metric.

A further issue with these approaches is that the compliance of the
joints in many commonly used robotic hands is also non-linear. A joint
powered by a highly geared motor will passively resist very large
external torques (up to the mechanical failure of the gears). Thus, if
the contact force on a link increases due to the external wrench, the
joint torque will passively increase to match. However, even if
contact force decreases, the joint torque can not decrease below the
level actively applied by the motor. This is a non-linear effect that
existing analysis methods can not account for.

\section{Grasp Model}

Due to the above limitations of linear models, we propose a model that accounts for non-linear effects due to the behavior of contact forces and non-backdrivable actuators. To capture the passive behavior of the system in response to external
disturbance, we (as others before ~\cite{HANAFUSA77,BICCHI93,BICCHI94,BICCHI95,PRATTICHIZZO97}) rely on computing virtual object movements in conjunction with virtual springs placed at the contacts between the rigid object and the hand mechanism. Unlike previous work however, we attempt to also capture
effects that are non-linear with respect to virtual object movement: joint
torques that can not decrease below the commanded levels (but can
increase if the joint does not backdrive), as well as contact forces
restricted to the inside of the (linearized) friction cone. 

\subsection{Friction Model}

In order to express (linearized) friction constraints at each contact, contact forces are expressed as linear combinations of the edges that define contact friction pyramids, and restricted to lie inside the pyramid:
\begin{eqnarray}
\bm{D} \bm{\beta} &=& \bm{c} \label{eq:betas}\\
\bm{F} \bm{\beta} &\leq& 0 \label{eq:friction}
\end{eqnarray}

Details on the construction of the linear force expression matrix
$\bm{D}$ and the friction matrix $\bm{F}$ can be found in the work of
Miller and Christensen~\cite{MILLER03B}. We note that, while the friction model is linear, frictional forces are not linearly related to virtual object movements (in contrast to the linear compliance model discussed in the previous section). In fact, friction forces are not related to virtual object motion at all. Instead, we propose an algorithm that searches for equilibrium contact forces everywhere inside the friction cones. In this study we
use the Point Contact With Friction model, approximating Coulomb friction between rigid bodies. However, the formulation is general
enough for other linearized models, such as the Soft Finger
Contact~\cite{CIOCARLIE07b}. 

\subsection{Compliance Model}

Assuming virtual springs placed at the contacts, the normal force at a contact $i$ is determined by the virtual relative motion between the object and the robot hand at that contact in the direction of the contact normal. This can be expressed in terms of virtual object displacements $\bm{x}$ and virtual joint movements $\bm{q}$. As it is only the relative motion at the contact in the direction of the contact normal we are interested in, 
a subscript $n$ denotes a normal component of a contact force or relative motion. For simplicity, we choose unity stiffness for the virtual contact spring ($k=1$).
\begin{equation}
\bm{c}_{i,n} = k(\bm{G}^T \bm{x} - \bm{J} \bm{q})_{i,n} \label{eq:spring}
\end{equation}

However, a contact may only apply positive force (it may only push, not pull). Hence, if the virtual object and joint movements are such that the virtual spring is extended from its rest position, the contact force must be zero. Thus, the virtual springs operate in two regimes:
\begin{enumerate}
  \item The object and hand are moving such as to compress the virtual spring at the contact. The contact force is positive ($\bm{c}_{i,n} \geq 0$) and the equality in (\ref{eq:spring}) holds.
  \item The object and hand are moving away from each other at the contact. The contact force is zero ($\bm{c}_{i,n} = 0$) and (\ref{eq:spring}) no longer holds: $\bm{c}_{i,n} - k(\bm{G}^T \bm{x} - \bm{J} \bm{q})_{i,n} \geq 0$.
\end{enumerate}

We can devise the following set of equations, which capture this behavior.
\begin{eqnarray}
\bm{c}_{i,n} &\geq& 0\\
\bm{c}_{i,n} - k(\bm{G}^T \bm{x} - \bm{J} \bm{q})_{i,n} &\geq& 0\\
\bm{c}_{i,n} \cdot (\bm{c}_{i,n} - k(\bm{G}^T \bm{x} - \bm{J} \bm{q})_{i,n}) &=& 0
\end{eqnarray}

This is a non-convex quadratic constraint and as such not readily solvable. (Note that if re-posed as a Linear Complementarity Problem it produces a non positive-definite matrix relating the vectors of unknowns). However, the same problem can be posed as a set of linear inequality constraints instead, which can be solved by a mixed-integer programming solver.
\begin{eqnarray}
\bm{c}_{i,n} &\geq& 0 \label{eq:springsMIPbegin}\\
\bm{c}_{i,n} &\leq& k_1 \cdot y_i\\
\bm{c}_{i,n} - k(\bm{G}^T \bm{x} - \bm{J} \bm{q})_{i,n} &\geq& 0\\
\bm{c}_{i,n} - k(\bm{G}^T \bm{x} - \bm{J} \bm{q})_{i,n} &\leq& k_2 \cdot (1-y_i) \label{eq:springsMIPend}
\end{eqnarray}

Each contact $i$ is assigned a binary variable $y_i \in \{0,1\}$ determining the regime, in which the virtual spring operates and hence if the normal force at that contact is equal to the force in the virtual spring (for positive spring forces) or zero. Constants $k_1$ and $k_2$ are virtual limits that have to be carefully chosen such that the magnitude of the expressions on the left-hand side never exceed them. However, they should not be chosen too large or the problem may become numerically ill-conditioned.

\subsection{Joint Model}

The mechanics of the hand place constraints on the virtual motion of the joints. To clarify this point, consider the equilibrium joint torque $\tau$, at which the system settles, and which may differ from the commanded joint torque $\tau_c$. At any joint $j$ the torque may exceed the commanded value, but only passively. In non-backdrivable hands this means the torque at a joint may only exceed its commanded level if the gearing between the motor and the joint is absorbing the additional torque. In consequence, a joint at which the torque exceeds the commanded torque may not display virtual motion. Similarly to the behavior of the virtual springs at the contacts, the relationship between joint torque and virtual joint motion exhibits two distinct regimes. Therefore - defining joint motion which closes the hand on the object as positive - this constraint can be expressed as another linear complementarity.

\begin{eqnarray}
q_j &\geq& 0\\
\tau_j - \tau_{c,j} &\geq& 0\\
q_j \cdot (\tau_j - \tau_{c,j}) &=& 0
\end{eqnarray}

Similarly to the linear complementarity describing normal contact forces this constraint can be posed as a set of linear inequalities.
\begin{eqnarray}
q_j &\geq& 0 \label{eq:torqueMIPbegin}\\
q_j &\leq& k_3 \cdot z_j \\
\tau_j - \tau_{c,j} &\geq& 0 \\
\tau_j - \tau_{c,j} &\leq& k_4 \cdot (1-z_j) \label{eq:torqueMIPend}
\end{eqnarray}

Each joint is assigned a binary variable $z_j$ that determines if the joint may move or is being passively loaded and hence stationary. Similarly to $k_1$ and $k_2$ the constants $k_3$ and $k_4$ are virtual limits and should be chosen with the same considerations in mind. 

\subsection{Underactuation Model}

The above joint model assumes a direct drive robotic hand kinematic, where an actuator command equates to an individual joint torque command. However, our framework is well suited to model hand kinematics, where the joint torques can be expressed as linear combinations of actuator forces. This includes underactuated designs with fewer actuators than degrees of freedom such as, for example, a tendon driven hand with fewer tendons than joints. This implies, that a tendon - and hence an actuator - can directly apply torques to multiple joints by means of a mechanical transmission. We thus define matrix $\bm{R}$, which maps from forces at the actuators \bm{$f$} to joint torques \bm{$\tau$}. Note, that its transpose maps from joint motion to the motion of the mechanical force transmission at the actuator.
\begin{equation}
\bm{R} \bm{f} = \bm{\tau} \label{eq:torqueRatio}
\end{equation}

Again, we assume the actuators to be non backdrivable and hence at an actuator $l$ we may see a force $f_l$ that exceeds the commanded value $f_{l,c}$ - and again, this can only occur passively. This means, that an actuator force can only exceed the commanded value, if the actuator is being backdriven, and hence the mechanical transmission does not exhibit any virtual motion. Defining transmission motion $\bm{R^T} \bm{q}$ that closes the hand around the object as positive, we can again express this constraint was a linear complementarity.
\begin{eqnarray}
(\bm{R^T}\bm{q})_l &\geq& 0\\
f_l - f_{c,l} &\geq& 0\\
(\bm{R^T}\bm{q})_l \cdot (f_l - f_{c,l}) &=& 0
\end{eqnarray}

Similarly to the previously described linear complementarities this constraint can be posed as a set of linear inequalities.
\begin{eqnarray}
(\bm{R^T}\bm{q})_l &\geq& 0 \label{eq:underactuationMIPbegin}\\
(\bm{R^T}\bm{q})_l &\leq& k_5 \cdot z_l \\
f_l - f_{c,l} &\geq& 0 \\
f_l - f_{c,l} &\leq& k_6 \cdot (1-z_j) \label{eq:underactuationMIPend}
\end{eqnarray}

Instead of a binary variable at each joint determining if it may move or is being loaded, we assign a binary variable to each actuator. This variable determines if the actuator - and hence the connected transmission - may move. The constants $k_5$ and $k_6$ are virtual limits just as constants $k_1$ through $k_4$.

This actuation model is a generalization of the previously introduced joint model and will reduce as such if the actuators control individual joint torques directly.

\section{Solution Method}

The computational price we pay for considering these non-linear effects is that virtual object movement is not directly determined by the
compliance-weighted inverse of the grasp map matrix; rather, it
becomes part of the complex mixed-integer problem we are trying to solve. In general, if a solution exists, there is an infinite number of solutions satisfying the constraints. The introduction of an optimization objective leads to a single solution. A physically well motivated choice of objective might be to minimize the energy stored in the virtual springs. We formulate a \textit{passive response problem} (or PRP) as outlined in Algorithm~\ref{alg:noniterative}.

\begin{algorithm}[!t]
\caption{}\label{alg:noniterative}
\begin{algorithmic}[0]
\State \textbf{Input:} $\bm{\tau}_c\ or\ \bm{f}_c$ - commanded joint torques/ actuator forces, $\bm{w}$ - applied wrench
\State \textbf{Output:} $\bm{c}$ - equilibrium contact forces
\Procedure{Passive Response Problem}{$\bm{\tau}_c, \bm{w}$} 
  \State \textbf{minimize:} $\bm{c}_n^T \bm{c}_n$ \Comment{energy stored in virtual springs}
  \State \textbf{subject to:} 
    \State \hspace{\algorithmicindent} $Eqs.\ (\ref{eq:wrench})\ \&\ (\ref{eq:tau})$ \Comment{equilibrium}
    \State \hspace{\algorithmicindent} $Eqs.\ (\ref{eq:betas})\ \&\ (\ref{eq:friction})$ \Comment{friction}
    \State \hspace{\algorithmicindent} $Eqs.\ (\ref{eq:springsMIPbegin})-(\ref{eq:springsMIPend})$  \Comment{compliance model}
    \State \hspace{\algorithmicindent} $Eqs.\ (\ref{eq:torqueMIPbegin})-(\ref{eq:torqueMIPend})\ or\ (\ref{eq:torqueRatio})\ \&\ (\ref{eq:underactuationMIPbegin})-(\ref{eq:underactuationMIPend})$ 
    \State \hspace{\algorithmicindent} \Comment{joint or underactuation model}
\State \textbf{return} $\bm{c}$ 
\EndProcedure
\end{algorithmic}
\end{algorithm}

\begin{figure*}[t!]
\centerline{\subfloat[PRP Solution (Alg.~\ref{alg:noniterative})]{\includegraphics[width=2.5in]{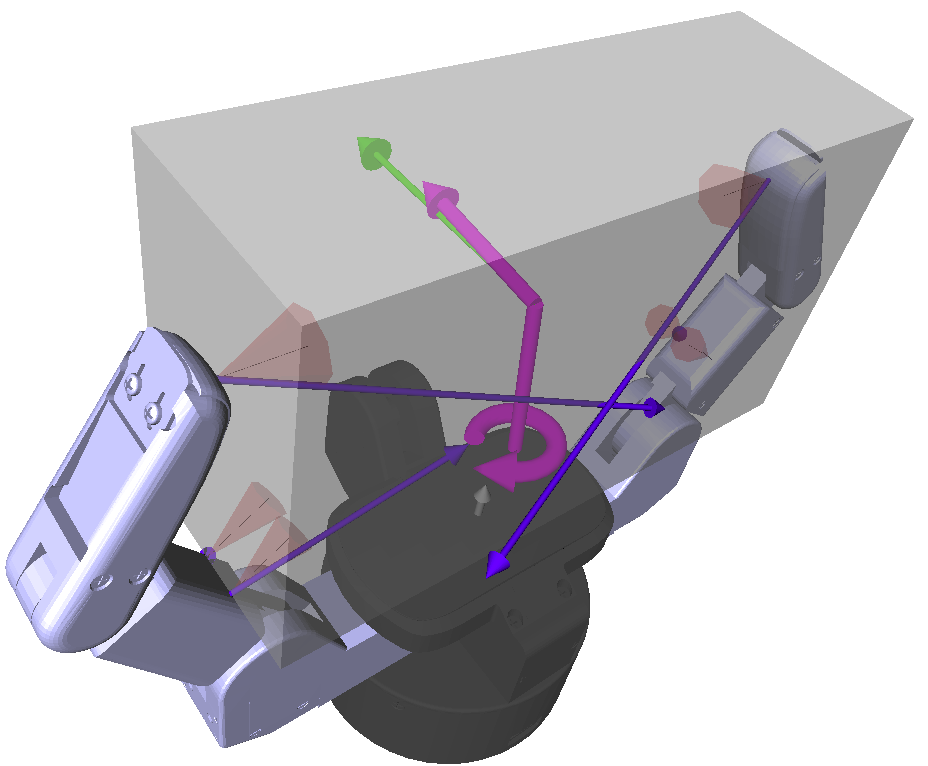}%
\label{fig:noniterative}}
\hfil
\subfloat[Iterative PRP Solution (Alg.~\ref{alg:loop})]{\includegraphics[width=2.5in]{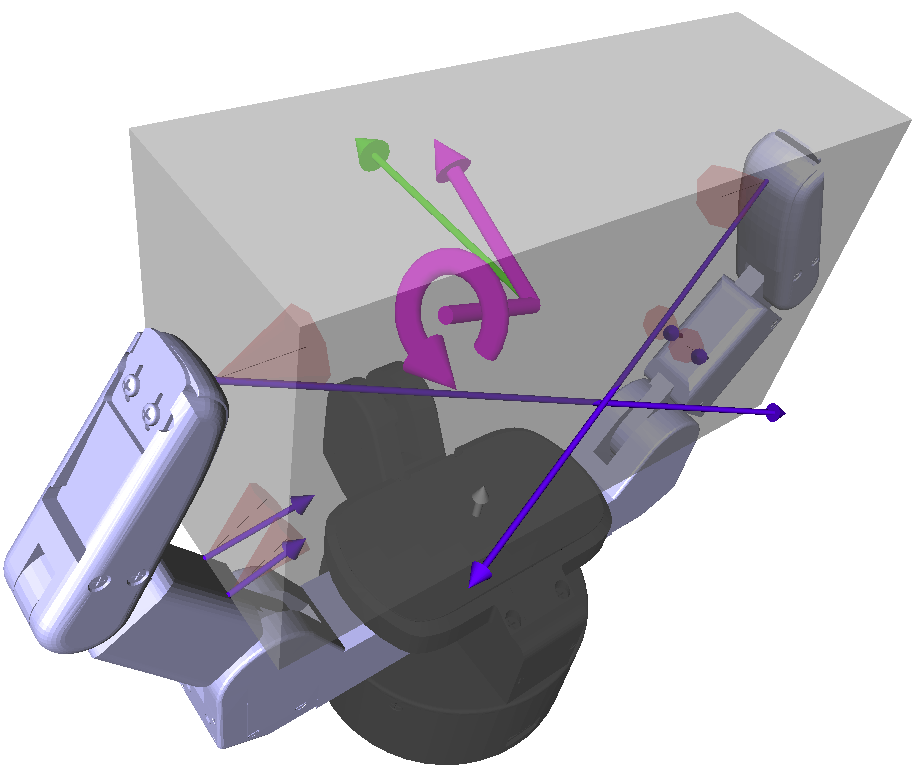}%
\label{fig:iterative}}}
\caption{Illustration of the shortcomings of directly solving the PRP problem defined above. A force normal to the closing plane of the fingers (illustrated by the green arrow) is applied to the object at its center of mass. The translational and rotational components of the resulting object movement are shown in violet. The PRP algorithm finds a way to wedge the object between the fingers by rotating the object in a way that would not occur in practice as it violates conservation of energy. This enables the solver to find ways to resist arbitrary wrenches. The iterative approach yields the natural finite resistance.}
\label{fig:wedge}
\end{figure*}

In certain circumstances, this formulation proves to be insufficient. The rigid, passively loaded fingers allow an optimization formulation with unconstrained object movement to ``wedge'' the object between contacts creating large contact forces. This allows the grasp to withstand very large applied wrenches by performing ``unnatural'' virtual displacements that satisfy all our constraints and lead to equilibrium, but violate the principle of conservation of energy: the energy stored in the virtual springs and the energy dissipated due to friction are greater than the work done by the externally applied wrench and the actuators (see Fig.~\ref{fig:noniterative}). Introducing a constraint that enforces energy conservation would solve this issue. However, friction is inherently non linear and calculation of the energy dissipated is thus non convex. Another approach would be to constrain the friction forces, which so far are only constrained by their respective cones. This could be done by enforcing the principle of least action: friction forces are to oppose motion and friction forces at sliding contacts must lie on the edge of the friction cone. This approach is non-linear as well.

As an exact treatment of the underlying physical laws in this formulation is non convex, we have devised an approximate iterative scheme that aims to eliminate unnatural object motion. We constrain the object movement such that motion is only allowed in the direction of the unbalanced wrench acting on the object: $\bm{x} = s\bm{w}, \quad s \in \mathbb{R}_{\ge 0}$. We remove the equilibrium constraint (\ref{eq:wrench}) and replace the objective of the optimization formulation such as to minimize the net resultant wrench $\bm{r}=\bm{w}+\bm{G}^T\bm{\beta}$ (the net sum of the applied wrench and contact forces) acting on the object. However, under this new constraint on virtual object motion, the solver will generally not be able to completely balance out the wrench and achieve equilibrium in a single step. Even after the optimization, some level of unbalanced wrench may remain. To eliminate it, we call the same optimization procedure in an iterative fashion, where, at each step we allow additional object movement in the direction of the unbalanced wrench $\bm{r}$ remaining from the previous call. For stability of the numerical scheme, we limit the step size by a parameter $\gamma$.
\begin{equation}
\bm{x}_{next} = \bm{x} + s\bm{r}, \quad 0 \leq s  \leq \gamma
\label{eq:movement}
\end{equation}

After each iteration, we check for convergence by comparing the incremental improvement to a threshold $\epsilon$. If the objective has converged to a sufficiently small net wrench (we chose $10^{-3}N$), we deem the grasp to be stable; otherwise, if the objective converges to a larger value, we deem the grasp unstable. Thus, we formulate a \textit{movement constrained passive response problem} as outlined in Algorithm~\ref{alg:iterative} to be solved iteratively as outlined in Algorithm~\ref{alg:loop}.

\begin{algorithm*}[!t]
\caption{}\label{alg:iterative}
\begin{algorithmic}[0]
\State \textbf{Input:} $\bm{\tau}_c\ or\ \bm{f}_c$ - commanded joint torques/ actuator forces, $\bm{w}$ - applied wrench, $\bm{x}$ - previous object displacement, $\bm{r}$ - previous net wrench
\State \textbf{Output:} $\bm{c}$ - contact forces, $\bm{x}_{next}$ - next step object displacement, $\bm{r}_{next}$ - next step net wrench
\Procedure{Movement Constrained PRP}{$\bm{\tau}_c, \bm{w}, \bm{x}, \bm{r}$}
  \State \textbf{minimize:} $\bm{r}^T_{next}\bm{r}_{next}$ \Comment{net wrench}
  \State \textbf{subject to:} 
    \State \hspace{\algorithmicindent} $Eq.\ (\ref{eq:tau})$ \Comment{torque/ contact force equilibrium}
    \State \hspace{\algorithmicindent} $Eqs.\ (\ref{eq:betas})\ \&\ (\ref{eq:friction})$ \Comment{friction}
    \State \hspace{\algorithmicindent} $Eqs.\ (\ref{eq:springsMIPbegin})-(\ref{eq:springsMIPend})$ \Comment{virtual springs complementarity}
    \State \hspace{\algorithmicindent} $Eqs.\ (\ref{eq:torqueMIPbegin})-(\ref{eq:torqueMIPend})\ or\ (\ref{eq:torqueRatio})\ \&\ (\ref{eq:underactuationMIPbegin})-(\ref{eq:underactuationMIPend})$ \Comment{joint or underactuation model}
    \State \hspace{\algorithmicindent} $Eq.\ (\ref{eq:movement})$ \Comment{object movement}
\State \textbf{return} $\bm{c}$, $\bm{x}_{next}, \bm{r}_{next}$ 
\EndProcedure
\end{algorithmic}
\end{algorithm*}

\begin{algorithm*}[!t]
\caption{}\label{alg:loop}
\begin{algorithmic}[0]
\State \textbf{Input:} $\bm{\tau}_c\ or\ \bm{f}_c$ - commanded joint torques/ actuator forces, $\bm{w}$ - applied wrench
\State \textbf{Output:} $\bm{c}$ - contact forces, $\bm{r}$ - net resultant
\Procedure{Iterative Passive Response Problem}{$\bm{\tau}_c, \bm{w}$}
  \State $\bm{x} = 0$
  \State $\bm{r} = \bm{w}$
  \Loop
    \State \textbf{($\bm{c},\bm{x}_{next},\bm{r}_{next}$)= Movement Constrained PRP($\bm{\tau}_c, \bm{w}, \bm{x}, \bm{r}$)} \Comment{Algorithm 2}
    \If{$norm(\bm{r}-\bm{r}_{next}) < \epsilon$} \Comment{Check if system has converged}
      \State \textbf{break}
    \EndIf
    \State $\bm{x} = \bm{x}_{next}$
    \State $\bm{r} = \bm{r}_{next}$  \EndLoop
  \State \textbf{return} $\bm{c}, \bm{r}$
\EndProcedure
\end{algorithmic}
\end{algorithm*}

The computation time of this process is directly related to the number of iterations required until convergence. A single iteration takes of the order of $10^{-2}$ to $10^{-1}$ seconds, depending on the complexity of the problem. Most problems converge within less than 50 iterations. All computations were performed on a commodity computer with a 2.80GHz Intel Core i7 processor.

We use this procedure to answer the question if a grasp, in which the joints are preloaded with a certain commanded torque can resist a given external wrench. In much of the analysis introduced in the next section we are interested in how the maximum external wrench, which a grasp can withstand depends on the direction of application. We approximate the maximum resistible wrench along a single direction using a binary search limited to 20 steps, which requires computation time of the order of tens of seconds. In general, investigating the magnitude of the maximum resistible wrench in every direction involves sampling wrenches in 6 dimensional space. Within our current framework this is prohibitively time consuming and hence we limit ourselves to sampling directions in 2 dimensions and then using the aforementioned binary search to find the maximum resistible wrench along those directions.

\section{Analysis and Results}

We illustrate the application of our method on three example
grasps using the Barrett hand and an underactuated gripper. We show 
force data collected by replicating the grasp on a real hand and 
testing resistance to external disturbances. We model the Barrett 
hand as having all non-backdrivable joints. Our
qualitative experience indicates that the finger flexion joints never
backdrive, while the spread angle joint backdrives under high
load. For simplicity we also do not use the breakaway feature of the
hand; our real instance of the hand also does not exhibit this
feature. We model the joints as rigidly coupled for motion, and assume
that all the torque supplied
by each finger motor is applied to the proximal joint. 

To measure the maximum force that a grasp can resist in a certain
direction, we manually apply a load to the grasped object using a
Spectra wire in series with a load cell (Futek, FSH00097). In order to
apply a pure force, the wire is connected such that the load direction
goes through the center of mass of the object. We increase the load
until the object starts moving, and take the largest magnitude
recorded by the load cell as the largest magnitude of the disturbance
the grasp can resist in the given direction.

\subsubsection{Case 1} 
\begin{figure}[!t]
\centering
\includegraphics[width=2.2in]{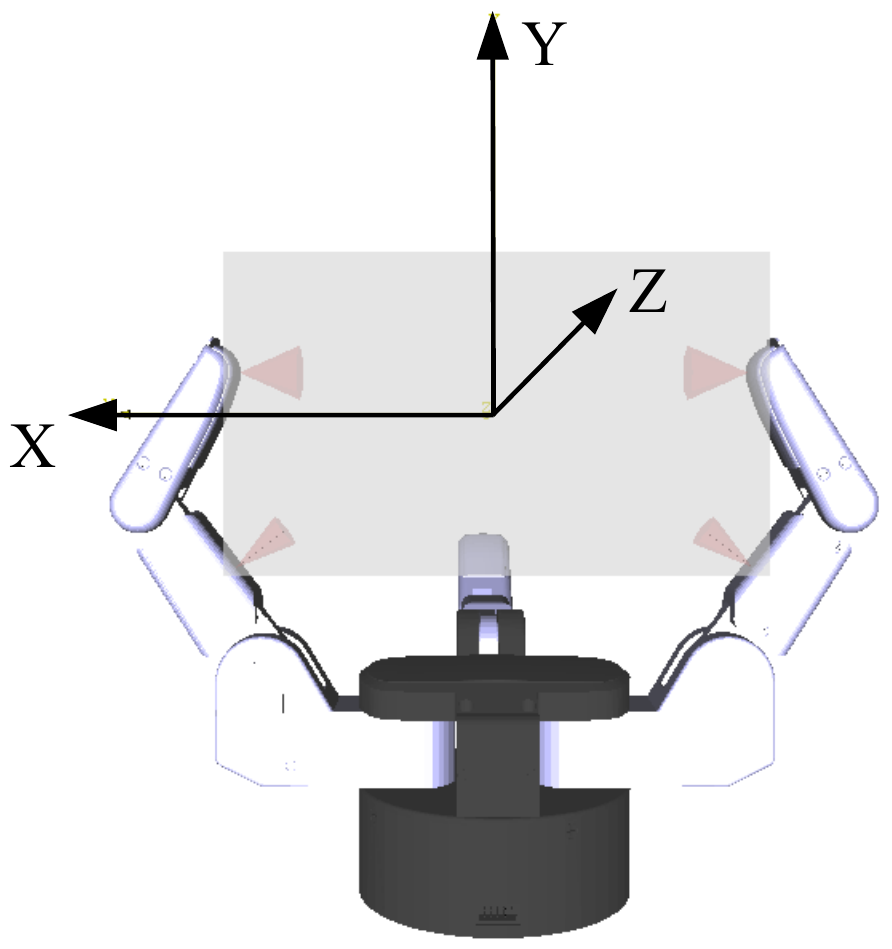}
\caption{Grasp example 1. Note that this is the same grasp we used in 
the Problem Statement section to explain the limitations of the linear 
compliance model.}
\label{fig:2dgrasp}
\end{figure}

We consider first the case illustrated in
Fig.~\ref{fig:2dgrasp}. This grasp can be treated as a 2D problem,
considering only forces in the grasp plane, simple enough to be
studied analytically, but still complex enough to give rise to
interesting interplay between the joints and contacts. Since our
simulation and analysis framework is built for 3D problems, we can
also study out-of plane forces and in-plane moments.
\begin{figure}[!t]
\centering
\includegraphics[width=3.27in]{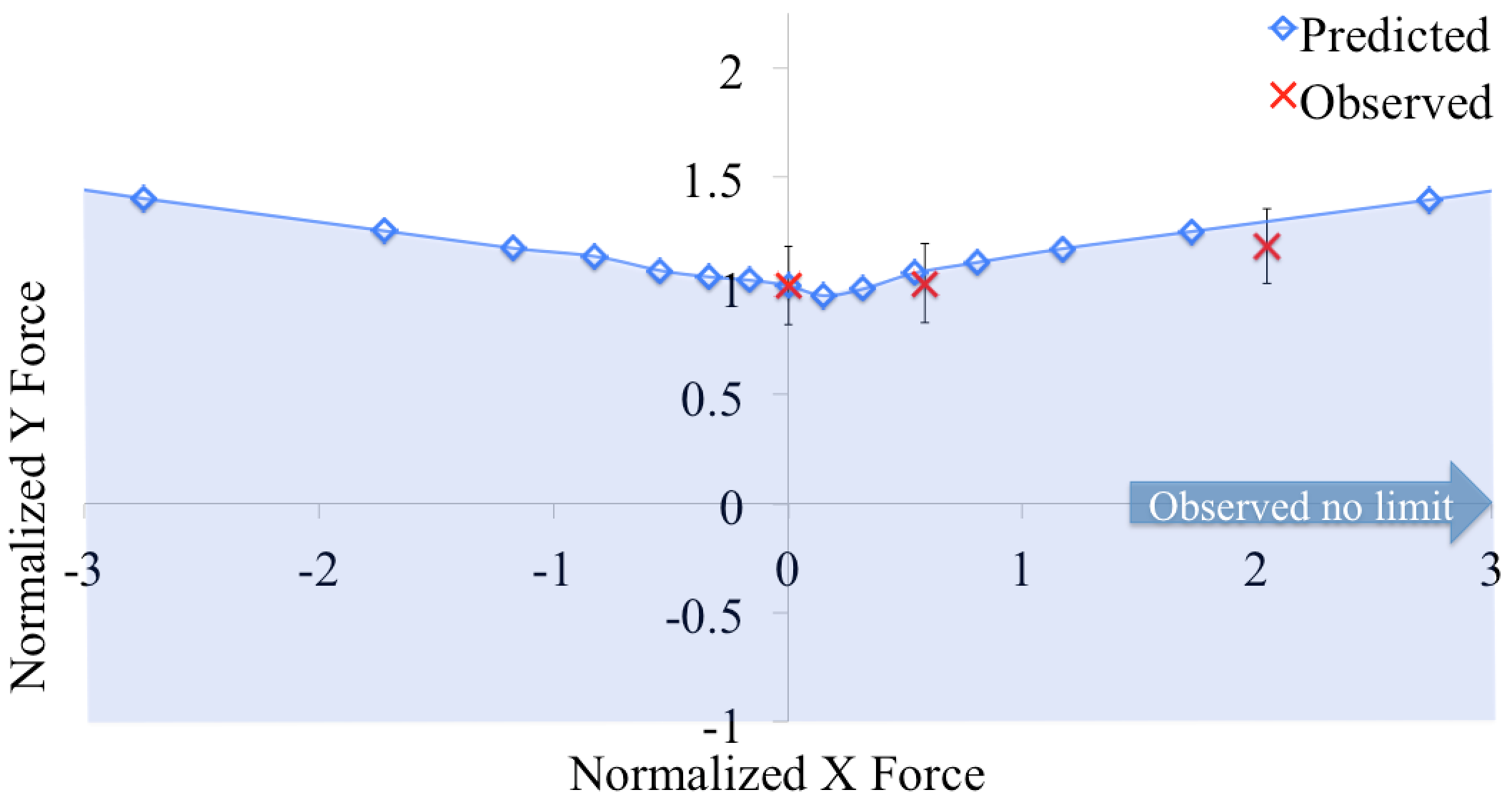}
\caption{Normalized forces in the XY plane that can be resisted by
  grasp example 1: observed by
  experiment (mean $\pm$ one standard deviation) and predicted by our 
  framework (normalized as explained in the text). In all directions
  falling below the blue line, the prediction framework hit the upper
  limit of the binary search (arbitrarily set to 1.0e3 N). Hence we 
  deem forces in the shaded area resistible. In the
  direction denoted by ``Observed no limit'', the grasp was not
  disturbed even when hitting the physical limit of the testing
  setup.}
\label{fig:2dgrasp_xy}
\end{figure}

Consider first the problem of resisting an external force applied to
the object CoM and oriented along the Y axis. This simple case already
illustrates the difference between active and passive resistance.
Resistance against a force oriented along positive Y requires active
torque applied at the joints in order to load the contacts and
generate friction. The force can be resisted only up to the limit
provided by the preload, along with the friction coefficient. If the
force is applied along negative Y, resistance happens passively,
provided through the contacts on the proximal link. Furthermore, this
resistant force does not require any kind of preload, and is infinite
(up to the breaking limit of the hand mechanism, which does not fall
within our scope here). 
\begin{figure}[!t]
\centering
\includegraphics[width=3.2in]{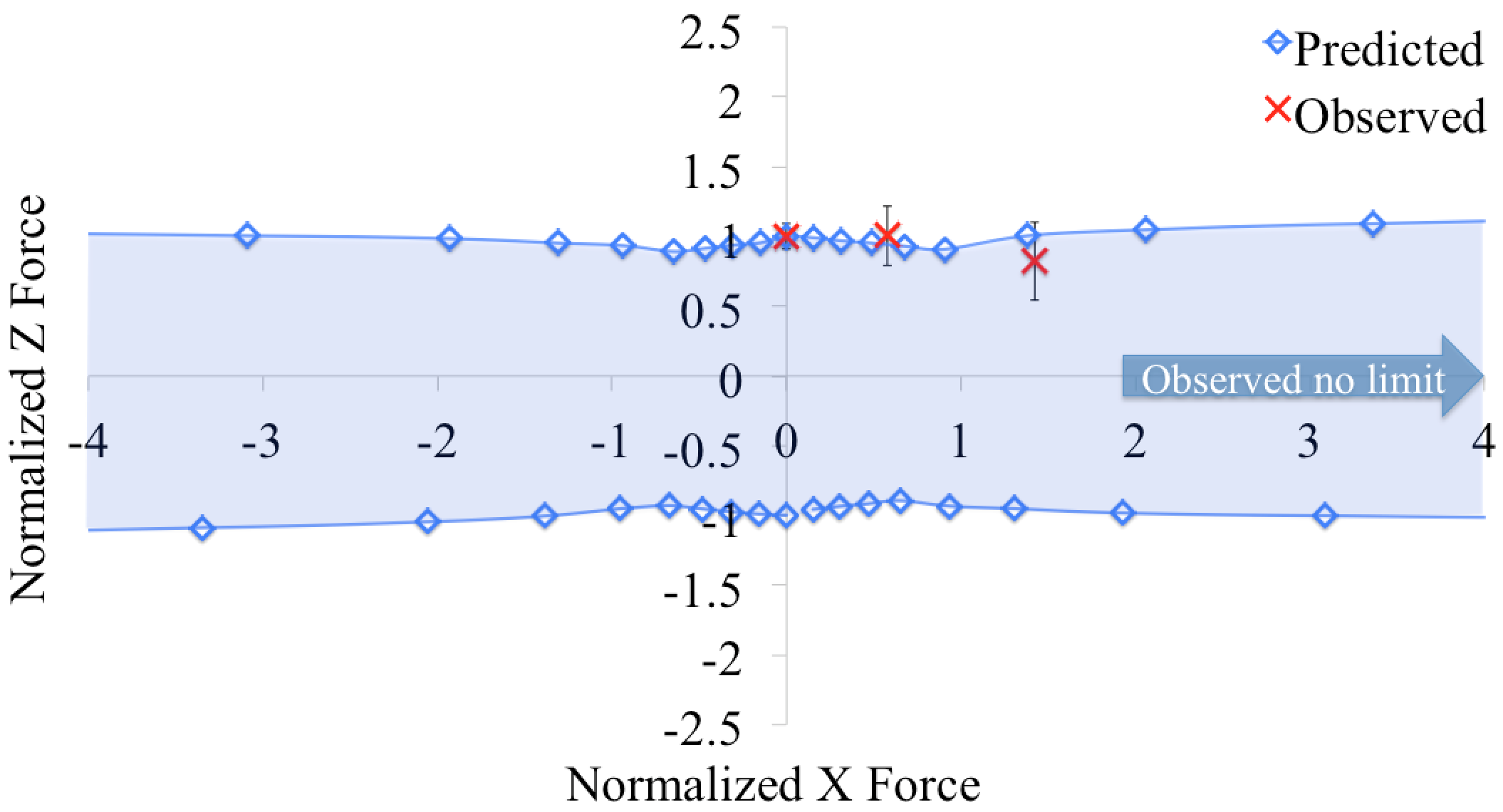}
\caption{Normalized forces in the XZ plane that can be resisted by
  grasp example 1: predicted by our framework, and observed by
  experiment. In all directions falling between the blue lines (shaded), the
  prediction framework hit the upper limit of the binary search
  (arbitrarily set to 1.0e3 N). In the direction denoted by ``Observed
  no limit'', the grasp was not disturbed even when hitting the
  physical limit of the testing setup.}
\label{fig:2dgrasp_xz}
\end{figure}

For an external force applied along the X axis, the problem is
symmetric between the positive and negative directions. Again, the
grasp can provide passive resistance, through a combination of forces
on the proximal and distal links. For the more general case of forces
applied in the XY plane, we again see a combination of active and passive
resistance effects. Intuitively, any force with a negative Y component
will be fully resisted passively. However, forces with a positive Y
component and non-zero X component can require both active and passive
responses. Fig.~\ref{fig:2dgrasp_xy} shows the forces that can be
resisted in the XY plane, both predicted by our framework and observed
by experiment. Note that our formulation predicts the distinction
between finite and infinite resistance directions, in contrast to the
results obtained using the linear compliance model.

For both real and predicted data, we normalize the force values by
dividing with the magnitude of the force obtained along the positive
direction of the Y axis (note thus that both predicted and
experimental lines cross the Y axis at y=1.0).  The plots should
therefore be used to compare trends rather than absolute values. We
use this normalization to account for the fact that the absolute
torque levels that the hand can produce, and which are needed by our
formulation in order to predict absolute force levels, can only be
estimated and no accurate data is available from the manufacturer. The
difficulty in obtaining accurate assessments of generated motor torque
generally limits the assessments we can make based on absolute force
values. However, if one knows the real magnitude of the maximum resistible
external force along any direction, in which this magnitude is finite, one could infer from these figures the real maximum
resistible wrenches in the other directions.

Moving outside of the grasp plane, Fig.~\ref{fig:2dgrasp_xz} shows
predicted and measured resistance to forces in the XZ plane. Again, we
notice that some forces can be resisted up to arbitrary magnitudes
thanks to passive effects, while others are limited by the actively
applied preload.

\subsubsection{Case 2} 
\begin{figure}[!t]
\includegraphics[width=3.32in]{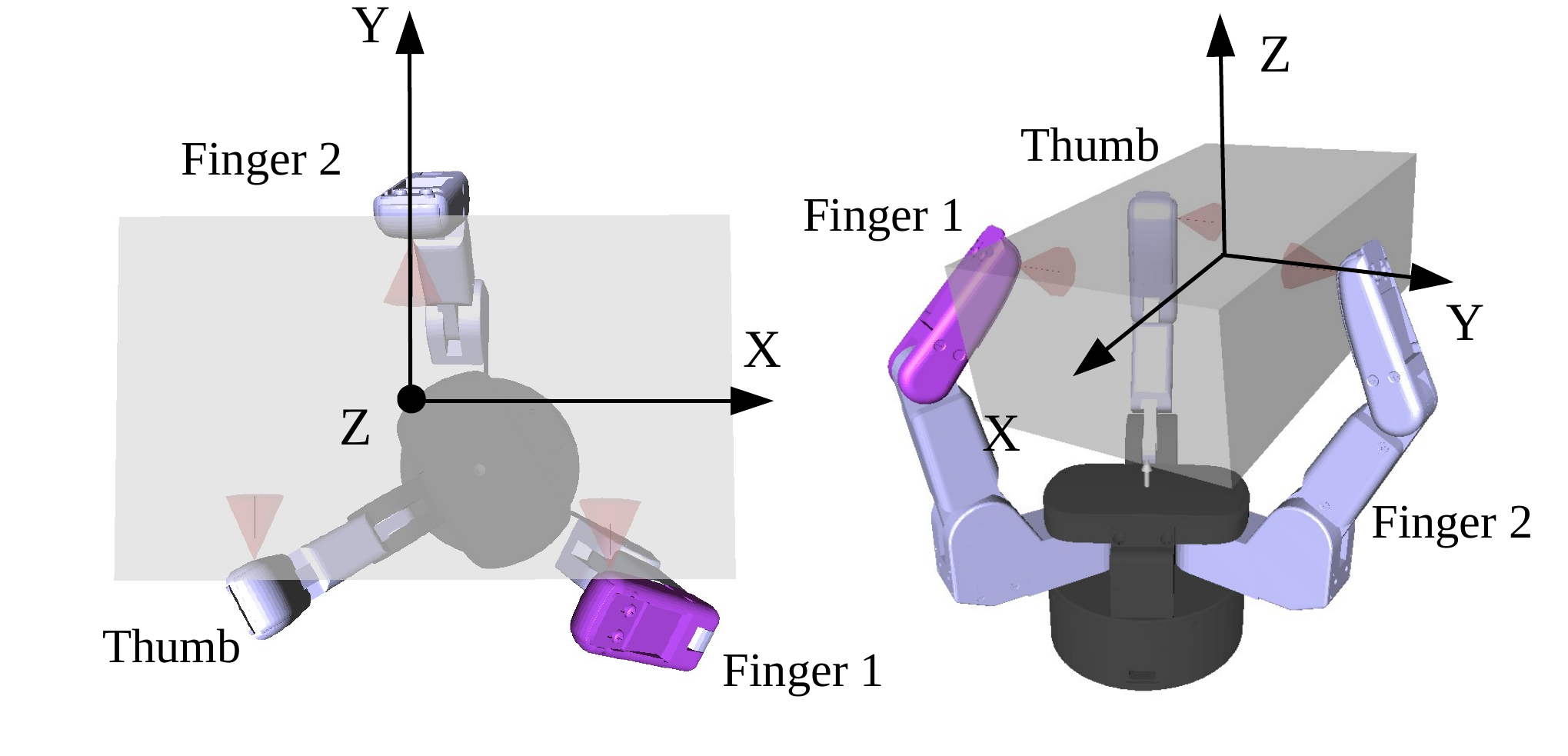}
\caption{Top and side views for grasp example 2 also indicating finger
  labels. Note that the spread angle degree of freedom of the Barrett
  hand changes the angle between finger 1 and finger 2; the thumb is
  only actuated in the flexion direction.}
\label{fig:3dgrasp}
\end{figure}

\begin{figure}[!t]
\centering
\includegraphics[width=2.8in]{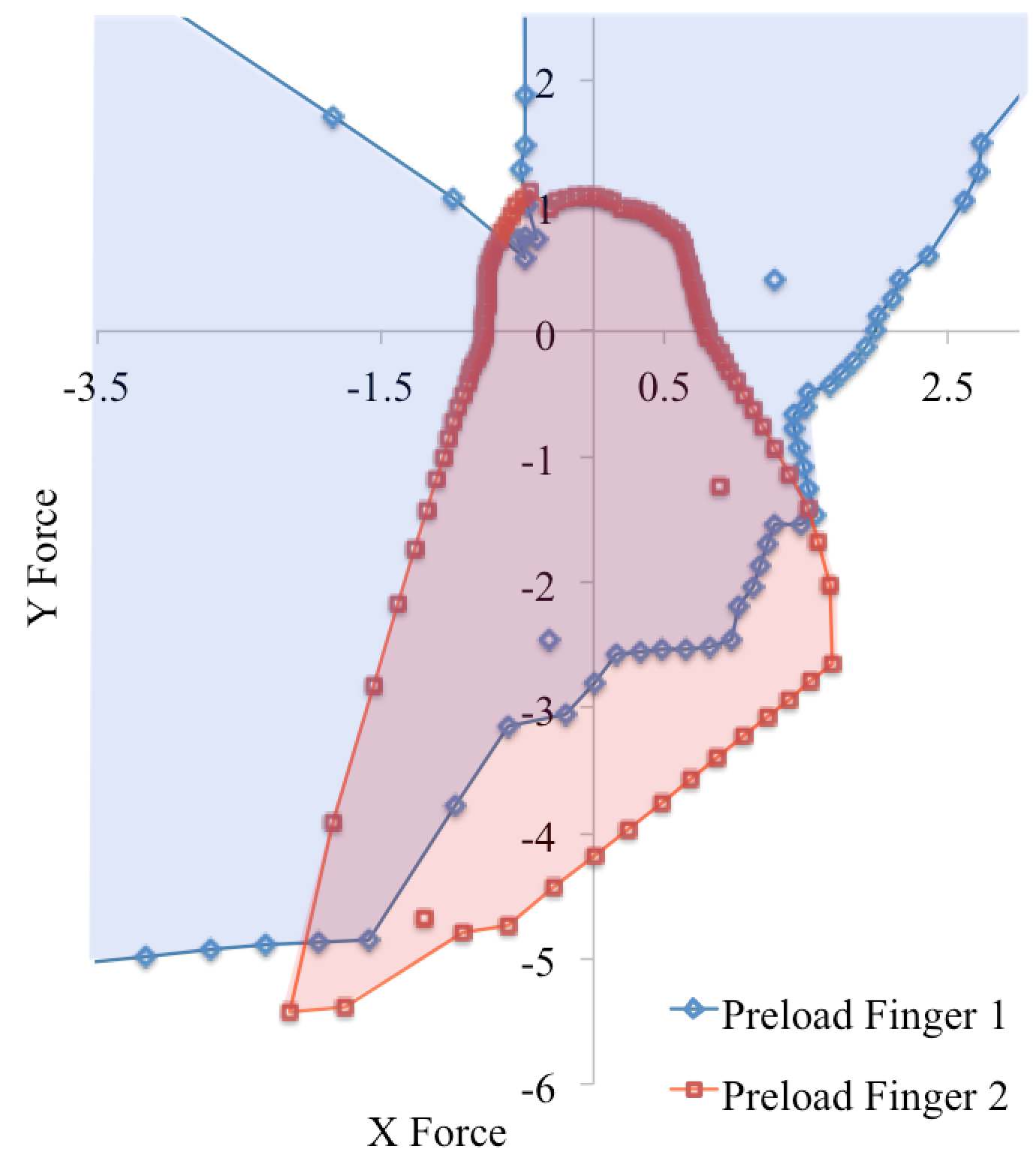}
\caption{Forces in an XY plane that can be resisted by grasp
  example 2 (shaded) as predicted by our framework, depending on which finger
  is preloaded. Note the four outlier results and that they have not been
  included in the determination of the regions of resistible forces. The 
  forces are normalized and hence dimensionless.}
\label{fig:outlier}
\end{figure}

One advantage of studying the effect of applied joint torques on grasp
stability is that it allows us to observe differences between
different ways of preloading the same grasp. For example, in the case
of the Barrett hand, choosing at which finger(s) to apply preload
torque can change the outcome of the grasp, even though there is no
change in the distribution of contacts. We illustrate this approach on the case shown in
Fig.~\ref{fig:3dgrasp}. Using our framework we can compute regions of 
resistible wrenches for two different preloads (see Fig.~\ref{fig:outlier}). 

We compare the ability of the grasp to resist
a disturbance applied along the X axis in the positive direction if
either finger 1 or finger 2 apply a preload torque to the grasp. Our
formulation predicts that by preloading finger 1 the grasp can resist
a disturbance that is 2.48 times higher in magnitude than if preloading
finger 2. Experimental data (detailed in Table~\ref{tab:3dgrasp})
indicates a ratio for the same disturbance direction of 2.23. The
variance in measurements again illustrates the difficulty of verifying
such simulation results with experimental data. Nevertheless,
experiments confirmed that preloading finger 1 is significantly better
for this case.
\begin{table*}[!t]
\renewcommand{\arraystretch}{1.3}
\caption{Predicted and measured resistance to force applied along the positive X axis in the grasp problem in Fig.~\ref{fig:3dgrasp}. Each row
  shows the results obtained if the preload is applied exclusively by
  finger 1 or finger 2 respectively. Experimental measurements were
  repeated 5 times for finger 1 (to account for the higher variance)
  and 3 times for finger 2. Predicted values are non dimensional, 
  and hence the ratio between the two preload cases is shown.}
\label{tab:3dgrasp}
\centering
\begin{tabular}{c|cccc|cc}
 & \multicolumn{4}{c}{Measured resistance} & \multicolumn{2}{c}{Predicted} \\
 & Values(N) & Avg.(N) & St. Dev. & Ratio & Value & Ratio\\\hline
&&&&&&\\[-3.5mm]\hline
&&&&&&\\[-2mm]
F1 load & 12.2, 10.8, 7.5, 7.9, 9.3 & 9.6 & 1.9 & 2.23 & 1.98 & 2.48\\
F2 load & 3.7, 4.1, 5.0 & 4.3 & 0.7 &    & 0.80 &
\end{tabular}
\end{table*}

This result can be explained by the fact that, somewhat
counter-intuitively, preloading finger 1 leads to larger contact forces
than preloading finger 2, even if the same torque is applied by
each motor. Due to the orientation of finger 1, the contact force on
finger 1 has a smaller moment arm around the finger flexion axes than
is the case for finger 2. Thus, if the same flexion torque is applied
in turn at each finger, the contact forces created by finger 1 will be
higher. In turn, due to passive reaction, this will lead to higher
contact forces on finger 2, even if finger 1 is the one being actively
loaded. Finally, these results hold if the spread degree of freedom is
rigid and does not backdrive; in fact, preloading finger 1 leads to a
much larger passive (reaction) torque on the spread degree of freedom
than when preloading finger 2.

Referring to Fig.~\ref{fig:outlier}, we note that actively preloading finger 
1 results in greater resistance only in some directions. There is much
structure to the prediction made by our framework that could be exploited
to make better decisions when preloading a grasp with some knowledge of the 
expected external wrenches. 

\subsubsection{Case 3}  
\begin{figure}[!t]
\centering
\includegraphics[width=3.0in]{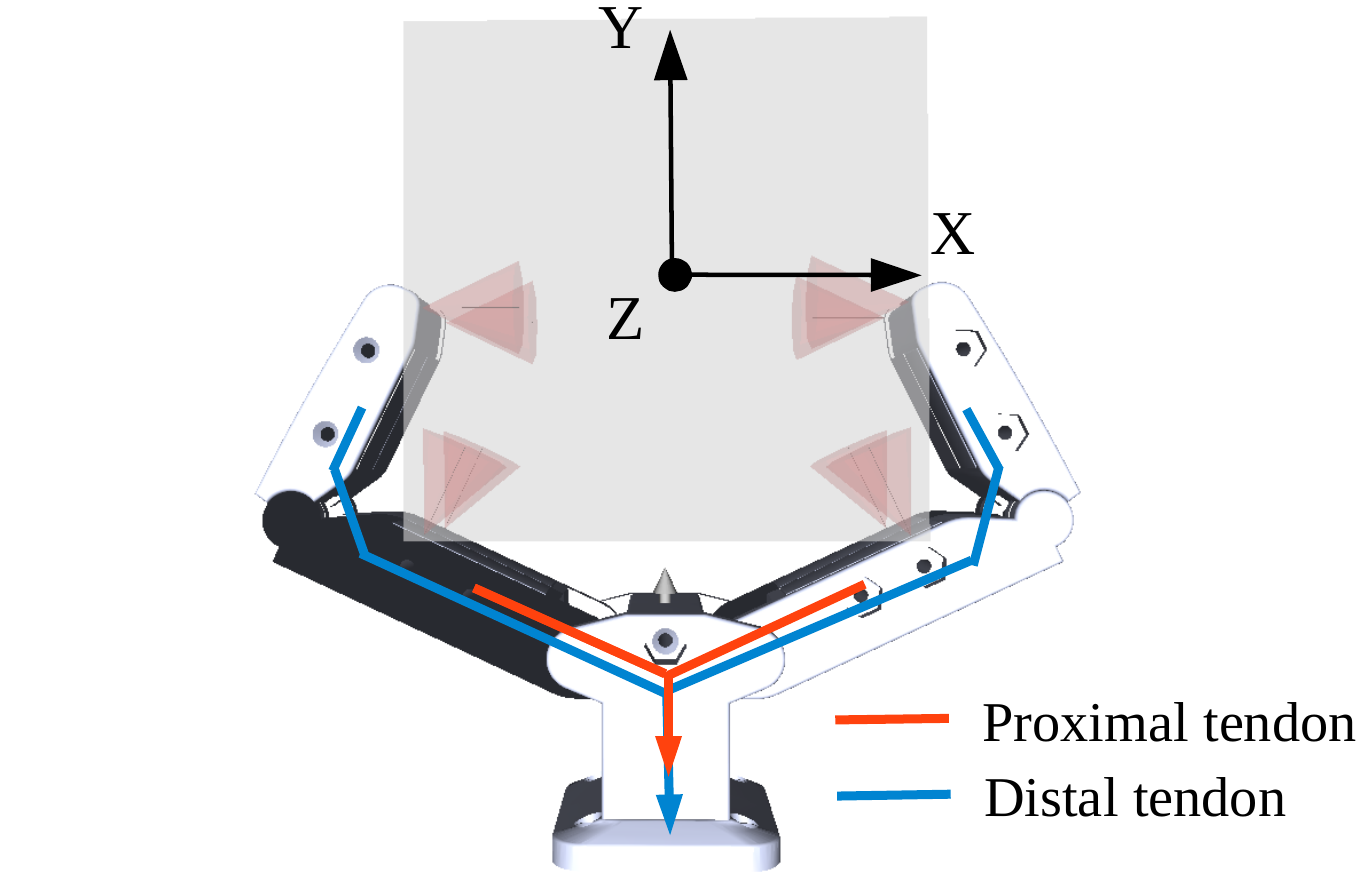}
\caption{Grasp example 3. Note that in terms of contact positions, this grasp is very similar to that in Fig.~\ref{fig:2dgrasp}. However, the kinematic differences cause these two grasps to behave very differently.}
\label{fig:underactuatedGrasp}
\end{figure}

We now apply our framework to a grasp with a two-fingered, tendon-driven 
underactuated gripper (see Fig.~\ref{fig:underactuatedGrasp}). The gripper 
has four degrees of freedom, but only two actuators driving a proximal and 
a distal tendon. The proximal tendon has a moment arm of 5mm around the 
proximal joints. The distal tendon has moment arms of 1.6mm and 5mm around 
the proximal and distal joints respectively. The actuators are non 
backdrivable and hence the tendons not only transmit actuation forces, 
they also provide kinematic constraints to the motion of the gripper's 
links. 

\begin{figure}[!t]
\centering
\includegraphics[width=2.5in]{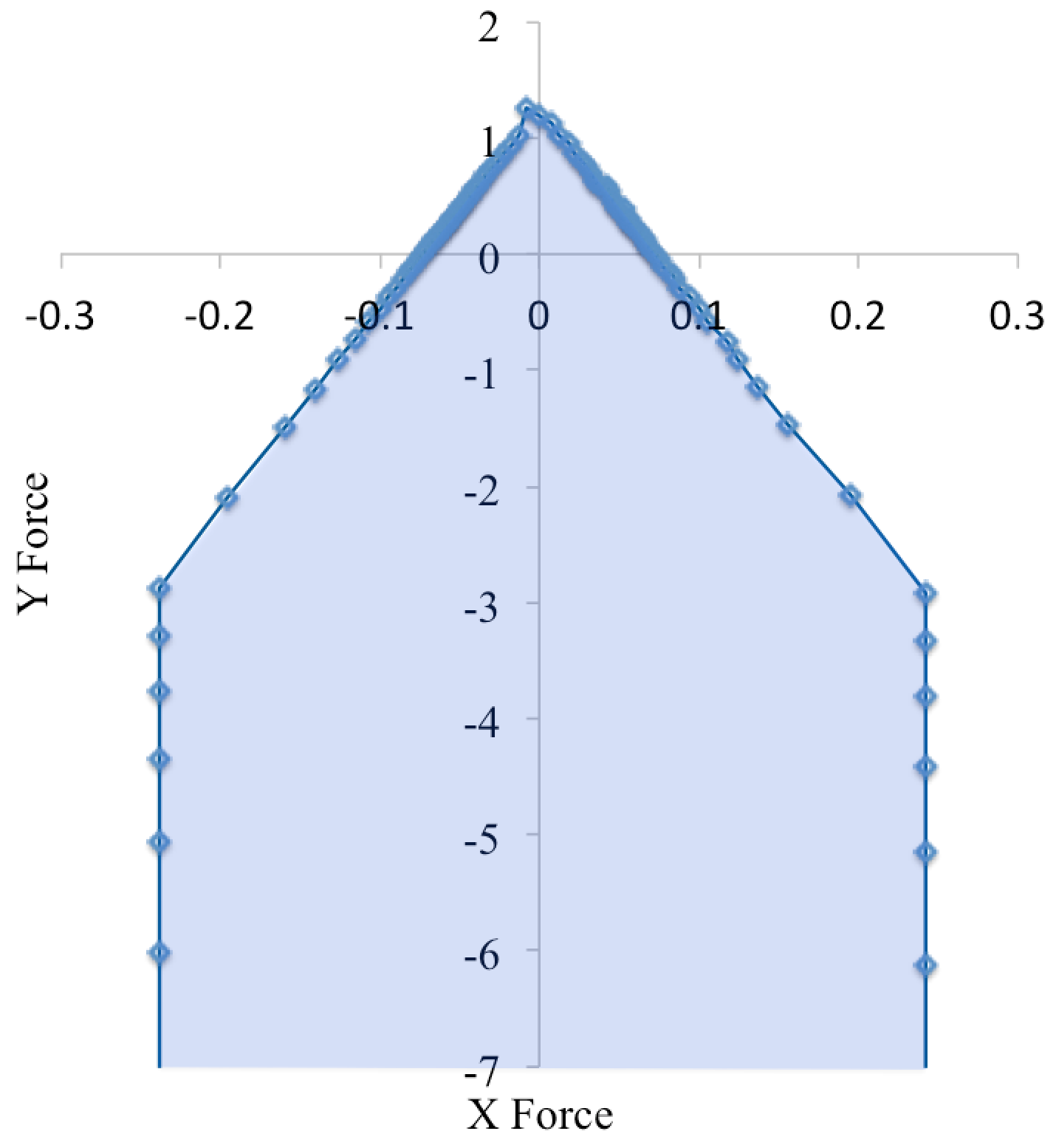}
\caption{Forces in an XY plane that can be resisted by grasp
  example 3 (shaded) as predicted by our framework. The 
  forces are normalized and hence dimensionless. Note the difference in scale on the X and Y axes.}
\label{fig:underactuated}
\end{figure}

As the tendons split and lead into both fingers, we assume that they are  
connected to the actuator by a differential. This introduces compliance to
the grasp: if one proximal joint closes by a certain amount, this
will allow the other proximal joint to open by a corresponding amount. 
This compliance means that the underactuated grasp in 
Fig.~\ref{fig:underactuatedGrasp} will behave fundamentally different
than the very similar grasp in Fig.~\ref{fig:2dgrasp}. To see this, 
consider the region of resistible wrenches in the XY plane 
(Figs.~\ref{fig:2dgrasp_xy}\ \&~\ref{fig:underactuated}). In both cases 
the object is gripped by two opposing fingers, however, while in the case 
of the Barrett hand the grasp could withstand forces pushing the object 
directly against a finger, our framework predicts this to be impossible 
in the case of the underactuated hand due to the compliance. 

We used our framework to apply two different preloads and analyzed the 
resistance of the resulting grasp to an externally applied wrench.
As our real underactuated hand does not contain a 
differential between the left and the right halves of the tendons, we can
only compare grasps to the simulation if the applied wrench 
does not cause any asymmetry in the grasp. Hence, we chose to apply a 
torque around the X axis. The two preload cases we considered are an active load on the proximal/ 
distal tendon only, leaving the distal/ proximal tendon to be loaded 
passively. For equal actuator force, our framework predicts, that a 
preload created by actively loading the distal tendon leads to almost 
twice as much resistance to torques applied in the direction of the Y 
axis than actively loading the proximal tendon only. 

Experimental verification of this prediction proved to be difficult, as 
results had high variance and application of a pure torque to the object 
along an axis that penetrates the distal links of the gripper was 
complicated. However, we mounted the gripper such that the grasp plane 
was in the horizontal and placed weights on the top end of the box 
object. We found the resistance to these applied wrenches indeed to be 
much higher when actively loading the distal tendon as opposed to the 
proximal.

\section{Discussion}

\subsubsection{Limitations} 
In a subset of cases, the solver reports maximum resistible wrenches 
with very different magnitude relative to neighboring
states. For example, in the grasp Case 2 from the previous section
(Fig.~\ref{fig:3dgrasp}), when computing resistance to
disturbances sampled from the XY plane
(Fig.~\ref{fig:outlier}), we obtain two outliers for each preload case 
that do not follow the trend of the surrounding points. These outliers are
quite rare and tend to fall within the area deemed to contain resistible wrenches
(shaded). These effects will require further investigation. 

Our iterative approach allows us to constrain virtual object movement
to the successive directions of unbalanced wrenches. However, such an
iterative approach is not guaranteed to converge, or to converge to
the physically meaningful state of the system. We would like to explore
other formulations and iterative schemes to better approximate the 
non linear and non convex physical laws governing the behavior of the 
grasp.

Our current real underactuated hand only allows experimental validation of a subset of our analysis results. We are working on designing a hand that we can use to further validate our 
framework and study the effects of underactuation on grasp stability. For 
instance, our framework predicts that wrench resistance is 
highly dependent on the torque ratios at the joints due to the kinematics
of the force transmission. We would like to experiment with a variety of
underactuated hands, with varying kinematic and actuation models, to investigate these
effects. 

Furthermore, we would like to analyze the effect of uncertainties (e.g. in 
exact contact location) on our model. We believe exploring the sensitivity 
of the model to such uncertainties may yield many valuable insights.

\subsubsection{Alternative Approaches}
As was described in the Problem Statement, a simpler alternative is to disregard non linear effects
with respect to virtual object movement, i.e. assume that the joints are fixed
(thus joint torque can both increase and decrease passively), and that
friction forces also behave in spring-like fashion. The price for this simplicity is, that the results may not be physically sound. 

At the other and of the spectrum, our iterative approach allowing
successive virtual object movements in the direction of the net
resultant wrench shares some of the features of a typical
time-stepping dynamics engine. One could therefore forgo the
quasi-static nature of our approach, assume that unbalanced wrenches
produce object acceleration or impulses, and perform time integration
to obtain new object poses. This approach can have additional
advantages: even an unstable grasp can eventually transform into a
stable one, as the object settles in the hand; a fully dynamic
simulation can capture such effects. However, in highly constrained
cases, such as grasps at or near equilibrium, any inaccuracy can lead
to the violation of interpenetration or joint constraints, in turn
requiring corrective penalty terms which add energy to the system. Our
quasi-static approach only attempts to determine if an equilibrium can
exist in the given state, and thus only reasons about virtual object
movements, without dynamic effects.

\section{Conclusions}
In this paper, we have introduced an algorithm that aims to answer
what we believe to be not only a meaningful theoretical question, but
also one with important practical applications: \textit{once a given
  joint preload has been achieved, can a grasp resist a given wrench
  passively, i.e. without any change in commanded joint torques?} In
the inner loop of a binary search, the same algorithm allows us to
determine the largest magnitude that can be resisted for a disturbance
along a given direction.

In the examples above we show how the actively set joint preload
combines with passive effects to provide resistance to external
wrenches; our algorithm captures these effects. Furthermore, we can
also compute how preloads set for some of the hand joints can cause
the other joints to load as well, and the combined effects can exceed
the intended or commanded torque levels. We can also study what subset
of the joints is preferable to load with the purpose of resisting
specific disturbances. Our grasp model captures well the effect 
compliance and underactuation have on grasp stability.

Our directional goal is to enable practitioners to choose grasps for a
dexterous robotic hand knowing that all disturbances they expect to
encounter will be resisted without further changes in the commands
sent to the motors. Such a method would have wide applicability, to
hands that are not equipped with tactile or proprioceptive sensors
(and thus unable to sense external disturbances) and can not
accurately control joint torques, but are still effective thanks to
passive resistance effects.

In its current form, the algorithm introduced here can answer ``point
queries'', for specific disturbances or disturbance
directions. However, its computational demands do not allow a large
number of such queries to be answered if a grasp is to be planned at
human-like speeds; furthermore, the high dimensionality of the
complete space of possible external wrenches generally prevents
sampling approaches. GWS-based approaches efficiently compute a global
measure of wrenches that can be resisted assuming perfect information
and controllability of contact forces. We believe passive resistance
has high practical importance for the types of hands mentioned above,
but no method is currently available to efficiently distill passive
resistance abilities into a single, global assessment of the grasp. We
will continue to explore this problem in future work.





\bibliographystyle{IEEEtran} \bibliography{bib/orthosis,bib/grasping,bib/thesis}

\begin{thebibliography}{10}
\providecommand{\url}[1]{#1}
\csname url@samestyle\endcsname
\providecommand{\newblock}{\relax}
\providecommand{\bibinfo}[2]{#2}
\providecommand{\BIBentrySTDinterwordspacing}{\spaceskip=0pt\relax}
\providecommand{\BIBentryALTinterwordstretchfactor}{4}
\providecommand{\BIBentryALTinterwordspacing}{\spaceskip=\fontdimen2\font plus
\BIBentryALTinterwordstretchfactor\fontdimen3\font minus
  \fontdimen4\font\relax}
\providecommand{\BIBforeignlanguage}[2]{{%
\expandafter\ifx\csname l@#1\endcsname\relax
\typeout{** WARNING: IEEEtran.bst: No hyphenation pattern has been}%
\typeout{** loaded for the language `#1'. Using the pattern for}%
\typeout{** the default language instead.}%
\else
\language=\csname l@#1\endcsname
\fi
#2}}
\providecommand{\BIBdecl}{\relax}
\BIBdecl

\bibitem{FERRARI92}
C.~Ferrari and J.~Canny, ``Planning optimal grasps,'' in \emph{IEEE
  International Conference on Robotics and Automation}, 1992, pp. 2290--2295.

\bibitem{handbook2008}
D.~Prattichizzo and J.~Trinkle, ``Grasping,'' \emph{Springer Handbook of
  Robotics}, 2008.

\bibitem{SALISBURY83}
J.~Salisbury and B.~Roth, ``Kinematic and force analysis of articulated
  mechanical hands,'' \emph{ASME Journal of Mechanisms, Transmissions, and
  Automation in Design}, vol. 105, pp. 35--41, 1983.

\bibitem{AICARDI96}
M.~Aicardi, G.~Casalino, and G.~Cannata, ``Contact force canonical
  decomposition and the role of internal forces in robust grasp planning
  problems,'' \emph{International Journal of Robotics Research}, vol.~15,
  no.~4, pp. 351--364, 1996.

\bibitem{KERR86}
J.~Kerr and B.~Roth, ``Analysis of multifingered hands,'' \emph{International
  Journal of Robotics Research}, vol.~4, no.~4, pp. 3--17, 1986.

\bibitem{YOSHIKAWA91}
T.~Yoshikawa and K.~Nagai, ``Manipulating and grasping forces in manipulation
  by multifingered robot hands,'' \emph{IEEE Transactions on Robotics and
  Automation}, vol.~7, no.~1, pp. 67--77, 1991.

\bibitem{BICCHI93}
A.~Bicchi, ``Force distribution in multiple whole-limb manipulation,'' in
  \emph{Robotics and Automation, 1993. Proceedings., 1993 IEEE International
  Conference on}.\hskip 1em plus 0.5em minus 0.4em\relax IEEE, 1993, pp.
  196--201.

\bibitem{BICCHI94}
------, ``On the problem of decomposing grasp and manipulation forces in
  multiple whole-limb manipulation,'' \emph{Robotics and Autonomous Systems},
  vol.~13, no.~2, pp. 127--147, 1994.

\bibitem{BICCHI95}
------, ``On the closure properties of robotic grasping,'' \emph{The
  International Journal of Robotics Research}, vol.~14, no.~4, pp. 319--334,
  1995.

\bibitem{PRATTICHIZZO97}
\BIBentryALTinterwordspacing
D.~Prattichizzo, J.~K. Salisbury, and A.~Bicchi, \emph{contact and grasp
  robustness measures: Analysis and experiments}.\hskip 1em plus 0.5em minus
  0.4em\relax Berlin, Heidelberg: Springer Berlin Heidelberg, 1997, pp. 83--90.
  [Online]. Available: \url{http://dx.doi.org/10.1007/BFb0035199}
\BIBentrySTDinterwordspacing

\bibitem{PRATTICHIZZO13}
D.~Prattichizzo, M.~Malvezzi, M.~Gabiccini, and A.~Bicchi, ``On motion and
  force controllability of precision grasps with hands actuated by soft
  synergies,'' \emph{IEEE Transactions on Robotics}, vol.~29, no.~6, 2013.

\bibitem{YOSHIKAWA96}
T.~Yoshikawa, ``Passive and active closures by constraining mechanisms,'' in
  \emph{IEEE International Conference on Robotics and Automation}, vol.~2,
  1996, pp. 1477--1484.

\bibitem{MELCHIORRI97}
C.~Melchiorri, ``Multiple whole-limb manipulation: An analysis in the force
  domain,'' \emph{Robotics and Autonomous Systems}, vol.~20, no.~1, pp. 15--38,
  1997.

\bibitem{RIMON16}
J.~Burdick and E.~Rimon, ``Wrench resistant multi-finger hand mechanisms,'' in
  \emph{International Conference on Robotics and Automation}, 2016.

\bibitem{CLOUTIER18}
A.~Cloutier and J.~Yang, ``Grasping force optimization approaches for
  anthropomorphic hands,'' \emph{ASME Journal of Mechanisms and Robotics},
  vol.~10, 2018.

\bibitem{CUTKOSKY_COMPLIANCE}
M.~R. Cutkosky and I.~Kao, ``Computing and controlling the compliance of a
  robotic hand,'' \emph{IEEE Transactions on Robotics and Automation}, vol.~5,
  no.~2, 1989.

\bibitem{MALVEZZI13}
M.~Malvezzi and D.~Prattichizzo, ``Evaluation of grasp stiffness in
  underactuated compliant hands,'' in \emph{IEEE International Conference on
  Robotics and Automation}, 2013, pp. 2074--2079.

\bibitem{HANAFUSA77}
H.~Hanafusa and I.~Asada, ``Stable prehension by a robot hand with elastic
  fingers,'' in \emph{Proc. of the 7th ISIR, Tokyo}, 1977.

\bibitem{MILLER03B}
A.~Miller and H.~Christensen, ``Implementation of multi-rigid-body dynamics
  within a robotic grasping simulator,'' in \emph{IEEE lntl. Conference on
  Robotics and Automation}, 2003, pp. 2262--2268.

\bibitem{CIOCARLIE07b}
M.~Ciocarlie, C.~Lackner, and P.~Allen, ``Soft finger model with adaptive
  contact geometry for grasping and manipulation tasks,'' in \emph{Joint
  Eurohaptics Conference and IEEE Symp. on Haptic Interfaces}, 2007, pp.
  219--224.

\end{thebibliography}
%

%

\begin{IEEEbiography}[{\includegraphics[width=1in,height=1.25in,clip,keepaspectratio]{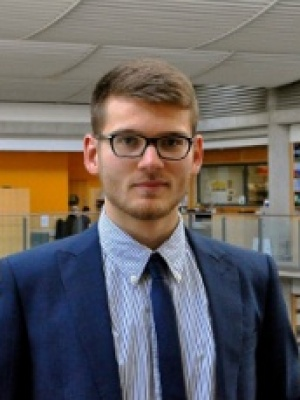}}]{Maximilian Haas-Heger}
Maximilian Haas-Heger received the MEng degree in Aeronautical Engineering from Imperial College London in 2015. Since 2015, he is a PhD candidate in the Robotic Manipulation and Mobility Lab at Columbia University in New York. His research focuses on the theoretical foundations of robotic grasping, specifically applied to the development of new grasp quality metrics. 
\end{IEEEbiography}

\begin{IEEEbiography}[{\includegraphics[width=1in,height=1.25in,clip,keepaspectratio]{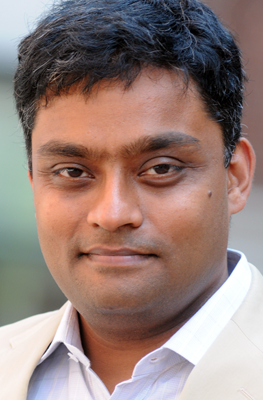}}]{Garud Iyengar}
Professor Garud Iyengar joined Columbia University’s Industrial Engineering and Operations Research Department in 1998. Professor Garud Iyengar’s research interests include convex optimization, robust optimization, queuing networks, combinatorial optimization, mathematical and computational finance, communication and information theory. He was elected as chairman of the IEOR Department on July 2013.
\end{IEEEbiography}

\begin{IEEEbiography}[{\includegraphics[width=1in,height=1.25in,clip,keepaspectratio]{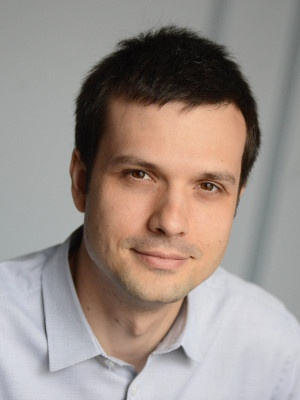}}]{Matei Ciocarlie}
Matei Ciocarlie is an Assistant Professor in the Mechanical Engineering Department at Columbia University, with affiliated appointments in Computer Science and the Data Science Institute. Before joining the faculty at Columbia, Matei was a Research Scientist and then Group Manager at Willow Garage, Inc., a privately funded Silicon Valley robotics research lab, and then a Senior Research Scientist at Google, Inc. Matei's current work focuses on robot motor control, mechanism and sensor design, planning and learning, all aiming to demonstrate complex motor skills such as dexterous manipulation.
\end{IEEEbiography}




\end{document}